\documentclass[11pt,a4paper]{article}
\usepackage[hyperref]{acl2020}
\usepackage{times}
\usepackage{latexsym}
\usepackage{balance}

\usepackage{microtype}

\aclfinalcopy 
\def\aclpaperid{22} 

\title{Solving Arithmetic Word Problems with Transformers and Preprocessing of Problem Text}

\author{Kaden Griffith and Jugal Kalita \\
  University of Colorado \\
  1420 Austin Bluffs Parkway \\
  Colorado Springs CO 80918 \\
  \texttt{kadengriffith@gmail.com and jkalita@uccs.edu}}

\date{12/07/2020}

\begin{document}

    \maketitle

    \begin{abstract}
        This paper outlines the use of Transformer networks trained to translate math word problems to equivalent arithmetic expressions in infix, prefix, and postfix notations. 
        We compare results produced by many neural configurations and find that most configurations outperform previously reported approaches on three of four datasets with significant increases in accuracy of over 20 percentage points. 
        The best neural approaches boost accuracy by 30\% when compared to the previous state-of-the-art on some datasets.
    \end{abstract}

    \section{Introduction}
        Students are exposed to simple arithmetic word problems starting in elementary school, and most become proficient in solving them at a young age. 
        However, it has been challenging to write programs to solve such elementary school level problems well.
        Even simple word problems, consisting of only a few sentences, can be challenging to understand for an automated system.

        Solving a math word problem (MWP) starts with one or more sentences describing a transactional situation.
        The sentences are usually processed to produce an arithmetic expression.
        These expressions then may be evaluated to yield a numerical value as an answer to the MWP.

        Recent neural approaches to solving arithmetic word problems have used various flavors of recurrent neural networks (RNN) and reinforcement learning.
        However, such methods have had difficulty achieving a high level of generalization.
        Often, systems extract the relevant numbers successfully but misplace them in the generated expressions.
        They also get the arithmetic operations wrong. 
        The use of infix notation also requires pairs of parentheses to be placed and balanced correctly, bracketing the right numbers. 
        There have been problems with parentheses placement.
        
        \begin{figure}
            \caption{Possible generated expressions for an MWP.}
            \label{figure:misrepresentations}
            \centering
            \begin{tabular}{p{0.94\linewidth}}
            \hline
            { \bf Question: } \\
            At the fair Adam bought 13 tickets. After riding the ferris wheel he had 4 tickets left. If each ticket cost 9 dollars, how much money did Adam spend riding the ferris wheel? \\ [.05in]
            { \bf Some possible expressions that can be produced: } \\
            {\small $(13 - 4) * 9, 9 * 13 - 4, 5 * 13 - 4, 13 - 4 * 9, 13 - (4 * 9),$ }
            {\small $(9 * 13 - 4), (9) * 13 -4, (9) * 13 - (4)$, etc. } \\
            \hline
            \end{tabular}
        \end{figure} 
        To start, correctly extracting the numbers in the problem is necessary.
        Figure \ref{figure:misrepresentations} gives examples of some infix representations that a machine learning solver can potentially produce from a simple word problem using the correct numbers.
        Of the expressions shown, only the first one is correct.
        The use of infix notation may itself be a part of the problem because it requires the generation of additional characters, the open and closed parentheses, which must be placed and balanced correctly.

        The actual numbers appearing in MWPs vary widely from problem to problem.
        Real numbers take any conceivable value, making it almost impossible for a neural network to learn representations for them.
        As a result, trained programs sometimes generate expressions that have seemingly random numbers.
        For example, in some runs, a trained program could generate a potentially inexplicable expression such as $(25.01 - 4) * 9$ for the problem given in Figure \ref{figure:misrepresentations}, with one or more numbers not in the problem sentences.
        To obviate this issue, we replace the numbers in the problem statement with generic tags like $\rm \langle i \rangle$, $\rm \langle q \rangle$, and $\rm \langle x \rangle$ and save their values as a preprocessing step.
        This approach does not take away from the generality of the solution but suppresses fertility in number generation leading to the introduction of numbers not present in the question sentences.
        We extend the preprocessing methods to ease the understanding through simple expanding and filtering algorithms.
        For example, some keywords within sentences are likely to cause the choice of operators for us as humans.
        By focusing on these terms in the context of a word problem, we hypothesize that our neural approach will improve further.

        In this paper, we use the Transformer model \cite{vaswani2017attention} to solve arithmetic word problems as a particular case of machine translation from text to the language of arithmetic expressions.
        Transformers in various configurations have become a staple of NLP in the past three years.
        We do not augment the neural architectures with external modules such as parse trees or deep reinforcement learning.
        We compare performance on four individual datasets.
        In particular, we show that our translation-based approach outperforms state-of-the-art results reported by \cite{wang2018mathdqn,hosseini2014learning,kushman2014learning,roy2015reasoning,robaidek2018data} by a large margin on three of four datasets tested.
        On average, our best neural architecture outperforms previous results by almost 10\%, although our approach is conceptually more straightforward.
        
        We organize our paper as follows.
        The second section presents related work.
        Then, we discuss our approach.
        We follow by an analysis of baseline experimental results and compare them to those of other recent approaches.
        We then take our best performing model and train it with motivated preprocessing, discussing changes in performance.
        We follow with a discussion of our successes and shortcomings.
        Finally, we share our concluding thoughts and end with our direction for future work.

    \section{Related Work}
        Past strategies have used rules and templates to match sentences to arithmetic expressions.
        Some such approaches seemed to solve problems impressively within a narrow domain but performed poorly otherwise, lacking generality  \cite{bobrow1964natural,bakman2007robust,liguda2012modeling,shi2015automatically}.
        Kushman et al. \cite{kushman2014learning} used feature extraction and template-based categorization by representing equations as expression forests and finding a close match.
        Such methods required human intervention in the form of feature engineering and the development of templates and rules, which is not desirable for expandability and adaptability.
        Hosseini et al. \cite{hosseini2014learning} performed statistical similarity analysis to obtain acceptable results but did not perform well with texts that were dissimilar to training examples.

        Existing approaches have used various forms of auxiliary information.
        Hosseini et al. \cite{hosseini2014learning} used verb categorization to identify important mathematical cues and contexts.
        Mitra and Baral \cite{mitra2016learning} used predefined formulas to assist in matching.
        Koncel-Kedziorski et al. \cite{koncel2015parsing} parsed the input sentences, enumerated all parses, and learned to match, requiring expensive computations.
        Roy and Roth \cite{roy2017unit} performed searches for semantic trees over large spaces.

        Some recent approaches have transitioned to using neural networks.
        Semantic parsing takes advantage of RNN architectures to parse MWPs directly into equations, or expressions in a  math-specific language \cite{shi2015automatically,sun2019neural}.
        RNNs have shown promising results, but they have had difficulties balancing parentheses.
	    Sometimes RNN models incorrectly choose numbers when generating equations.
        Rehman et al. \cite{rehman2019automatically} used part-of-speech tagging and classification of equation templates to produce systems of equations from third-grade level MWPs.
        Most recently, Sun et al. \cite{sun2019neural} used a bi-directional LSTM architecture for math word problems.
        Huang et al. \cite{huang-etal-2018-neural} used a deep reinforcement learning model to achieve character placement in both seen and new equation templates.
        Wang et al. \cite{wang2018mathdqn} also used deep reinforcement learning.
        We take a similar approach to \cite{wang2019template} in preprocessing to prevent ambiguous expression representation.
        
        This paper builds upon \cite{kaden}, extending the capability of similar Transformer networks and solving some common issues found in translations.
        Here, we simplify the tagging technique used in \cite{kaden}, and apply preprocessing to enhance translations.
        
    \section{Approach}
        We view math word problem solving as a sequence-to-sequence translation problem.
        RNNs have excelled in sequence-to-sequence problems such as translation and question answering.
        The introduction of attention mechanisms has improved the performance of RNN models.
        Vaswani et al. \cite{vaswani2017attention} introduced the Transformer network, which uses stacks of attention layers instead of recurrence.
        Applications of Transformers have achieved state-of-the-art performance in many NLP tasks.
        We use this architecture to produce character sequences that are arithmetic expressions.
        The models we experiment with are easy and efficient to train, allowing us to test several configurations for a comprehensive comparison.
        We use several configurations of Transformer networks to learn the prefix, postfix, and infix notations of MWP equations independently.

        Prefix and postfix representations of equations do not contain parentheses, which has been a source of confusion in some approaches.
        If the learned target sequences are simple, with fewer characters to generate, it is less likely to make mistakes during generation.
        Simple targets also may help the learning of the model to be more robust.

    \subsection{Data}
        We work with four individual datasets.
        The datasets contain addition, subtraction, multiplication, and division word problems.
        \begin{enumerate}
            \item {\bf AI2} \cite{hosseini2014learning}.
            AI2 is a collection of 395 addition and subtraction problems containing numeric values, where some may not be relevant to the question.
            \item {\bf CC} \cite{roy2016solving}.
            The Common Core dataset contains 600 2-step  questions.
            The Cognitive Computation Group at the University of Pennsylvania\footnote{\url{https://cogcomp.seas.upenn.edu/page/demos/}} gathered these questions.
            \item {\bf IL} \cite{roy2015reasoning}.
            The Illinois dataset contains 562 1-step algebra word questions.
            The Cognitive Computation Group compiled these questions also.
            \item {\bf MAWPS} \cite{koncel2016mawps}.
            MAWPS is a relatively large collection, primarily from other MWP datasets.
            MAWPS includes problems found in AI2, CC, IL, and other sources.
            We use 2,373 of 3,915 MWPs from this set.
            The problems not used were more complex problems that generate systems of equations.
            We exclude such problems because generating systems of equations is not our focus.
        \end{enumerate}

        We take a randomly sampled 95\% of examples from each dataset for training.
        From each dataset, MWPs not included in training make up the testing data used when generating our results.
        Training and testing are repeated three times, and reported results are an average of the three outcomes.

    \subsection{Representation Conversion}
        We take a simple approach to convert infix expressions found in the MWPs to the other two representations.
        Two stacks are filled by iterating through string characters, one with operators found in the equation and the other with the operands.
        From these stacks, we form a binary tree structure.
        Traversing an expression tree in preorder results in a prefix conversion.
        Post-order traversal gives us a postfix expression.
        We create three versions of our training and testing data to correspond to each type of expression.
        By training on different representations, we expect our test results to change.

    \subsection{Metric Used}
        We calculate the reported results, here and in later sections as:
        \begin{equation}
            \label{eq:averageAccuracy}
            model_{avg} = \frac{1}{R}\displaystyle\sum_{r=1}^{R} \bigg( \frac{1}{N}\displaystyle\sum_{n=1}^{N} \frac{C \in D_{n}}{P \in D_{n}} \bigg)
        \end{equation}
        where $R$ is the number of test repetitions, which is 3;
        $N$ is the number of test datasets, which is 4;
        $P$ is the number of MWPs;
        $C$ is the number of correct equation translations, and
        $D_{n}$ is the $n$th dataset.
        
        
    \section{Experiment 1: Search for High-Performing Models}
        The input sequence for a translation is a natural language specification of an arithmetic word problem.
        We encode the MWP questions and equations using the subword text encoder provided by the TensorFlow Datasets library.
        The output is an expression in prefix, infix, or postfix notation, which then can be manipulated further and solved to obtain a final answer.
        Each expression style corresponds to a model trained and tested separately on that specific style.
        For example, data in prefix will not intermix with data in postfix representation.

        All examples in the datasets contain numbers, some of which are unique or rare in the corpus.
        Rare terms are adverse for generalization since the network is unlikely to form good representations for them.
        As a remedy to this issue, our networks do not consider any relevant numbers during training.
        Before the networks attempt any translation, we preprocess each question and expression by a number mapping algorithm.
        We consider numbers found to be in word form also, such as ``forty-two" and ``dozen."
        We convert these words to numbers in all questions (e.g., ``forty-two" becomes ``42").
        Then by algorithm, we replace each numeric value with a corresponding identifier (e.g., $\langle j \rangle$, $\langle x \rangle$) and remember the necessary mapping.
        We expect that this approach may significantly improve how networks interpret each question.
        When translating, the numbers in the original question are tagged and cached.
        From the encoded English and tags, a predicted sequence resembling an expression presents itself as output.
        Since each network's learned output resembles an arithmetic expression (e.g., $\langle j \rangle + \langle x \rangle * \langle q \rangle$), we use the cached tag mapping to replace the tags with the corresponding numbers and return a final mathematical expression.

        We train and test three representation models: Prefix-Transformer, Postfix-Transformer, and Infix-Transformer.
        For each experiment, we use representation-specific Transformer architectures.
        Each model uses the Adam optimizer with $beta_1=0.95$ and $beta_2=0.99$ with a standard epsilon of $1 \times e^{-9}$.
        The learning rate is reduced automatically in each training session as the loss decreases.
        Throughout the training, each model respects a 10\% dropout rate.
        We employ a batch size of 128 for all training.
        Each model is trained on MWP data for 300 iterations before testing.
        The networks are trained on a machine using 4 Nvidia 2080 Ti graphics processing unit (GPU).

        We compare medium-sized, small, and minimal networks to show if a smaller network size can increase training and testing efficiency while retaining high accuracy.
        Networks over six layers have shown to be non-effective for this task.
        We tried many configurations of our network models but report results with only three configurations of Transformer.

        \begin{enumerate}
            \item[-] {\bf Transformer Type 1:}
            This network is a small to medium-sized network consisting of 4 Transformer layers.
            Each layer utilizes 8 attention heads with a depth of 512 and a feed-forward depth of 1024.
            \item[-] {\bf Transformer Type 2:}
            The second model is small in size, using 2 Transformer layers.
            The layers utilize 8 attention heads with a depth of 256 and a feed-forward depth of 1024.
            \item[-] {\bf Transformer Type 3:}
            The third type of model is minimal, using only 1 Transformer layer.
            This network utilizes 8 attention heads with a depth of 256 and a feed-forward depth of 512.
        \end{enumerate}

        \paragraph{Objective Function}
        We calculate the loss in training according to a mean of the sparse categorical cross-entropy formula.
        Sparse categorical cross-entropy \cite{de2005tutorial} is used for identifying classes from a feature set, assuming a large target classification set.
        The performance metric evaluates the produced class (predicted token) drawn from the translation classes (all vocabulary subword tokens).
        During each evaluation, target terms are masked, predicted, and then compared to the masked (known) value.
        We adjust the model's loss according to the mean of the translation accuracy after predicting every determined subword in a translation.
    
    \subsection{Experiment 1 Results}
        This experiment compares our networks to recent previous work.
        We count a given test score by a simple “correct versus incorrect” method. 
        The answer to an expression directly ties to all of the translation terms being correct, which is why we do not consider partial precision.
        We compare average accuracies over 3 test trials on different randomly sampled test sets from each MWP dataset. 
        This calculation more accurately depicts the generalization of our networks.
        
        We present the results of our various outcomes in Table 1. We compare the three representations of target equations and three architectures of the Transformer model in each test.
        \begin{table*}[!ht]
            \caption{Test results for Experiment 1 (* denotes averages on present values only). }
            \label{table:Experiment1Results}
            \centering
            \begin{tabular}{p{0.29\linewidth}p{0.11\linewidth}p{0.11\linewidth}p{0.11\linewidth}p{0.12\linewidth}p{0.1\linewidth}}
            \hline
            {\small \bf (Type) Model } & {\small \bf AI2 } & {\small \bf CC } & {\small \bf IL } & {\small \bf MAWPS } & {\small \bf Average } \\
            \hline
            {\small \cite{hosseini2014learning} } & {\small 77.7 } & {\small -- } & {\small -- } & {\small -- } & {\small $^*$77.7 } \\
            {\small \cite{kushman2014learning} } & {\small 64.0 } & {\small 73.7 } & {\small 2.3 } & {\small -- } & {\small $^*$46.7 } \\
            {\small \cite{roy2015reasoning} } & {\small -- } & {\small -- } & {\small 52.7 } & {\small -- } & {\small $^*$52.7 } \\
            {\small \cite{robaidek2018data} } & {\small -- } & {\small -- } & {\small -- } & {\small 62.8 } & {\small $^*$62.8 } \\
            {\small \cite{wang2018mathdqn} } & {\small \bf 78.5 } & {\small 75.5 } & {\small 73.3 } & {\small -- } & {\small $^*$75.4 } \\
            {\small (1) Prefix-Transformer } & {\small 71.9 } & {\small \bf 94.4 } & {\small \bf{95.2} } & {\small 83.4 } & {\small 86.3 } \\
            {\small (1) Postfix-Transformer } & {\small 73.7 } & {\small 81.1 } & {\small 92.9 } & {\small 75.7 } & {\small 80.8 } \\
            {\small (1) Infix-Transformer } & {\small 77.2 } & {\small 73.3 } & {\small 61.9 } & {\small 56.8 } & {\small 67.3 } \\
            {\small (2) Prefix-Transformer } & {\small 71.9 } & {\small \bf 94.4 } & {\small 94.1 } & {\small \bf 84.7 } & {\small 86.3 } \\
            {\small (2) Postfix-Transformer } & {\small 77.2 } & {\small \bf 94.4 } & {\small 94.1 } & {\small 83.1 } & {\small \bf 87.2 } \\
            {\small (2) Infix-Transformer } & {\small 77.2 } & {\small 76.7 } & {\small 66.7 } & {\small 61.5 } & {\small 70.5 } \\
            {\small (3) Prefix-Transformer } & {\small 71.9 } & {\small 93.3 } & {\small \bf{95.2} } & {\small 84.1 } & {\small 86.2 } \\
            {\small (3) Postfix-Transformer } & {\small 77.2 } & {\small 94.4 } & {\small 94.1 } & {\small 82.4 } & {\small 87.0 } \\
            {\small (3) Infix-Transformer } & {\small 77.2 } & {\small 76.7 } & {\small 66.7 } & {\small 62.4 } & {\small 70.7 } \\
            \hline
            \end{tabular}
        \end{table*}
        \begin{figure}
            \caption{Successful postfix translations.}
            \label{figure:successfulTranslations}
            \centering
            \begin{tabular}{p{0.9\linewidth}}
            \hline
            {\small \bf AI2 } \\
            {\small A spaceship traveled 0.5 light-year from earth to planet x and 0.1 light-year from planet x to planet y. Then it traveled 0.1 light-year from planet y back to Earth. How many light-years did the spaceship travel in all? } \\ [.05in]
            {\small \em Translation produced: } \\
            {\small 0.5 0.1 + 0.1 + } \\ [.1in]
            {\small \bf CC } \\
            {\small There were 16 friends playing a video game online when 7 players quit. If each player left had 8 lives, how many lives did they have total? } \\ [.05in]
            {\small \em Translation produced: } \\
            {\small 8 16 7 - * } \\ [.1in]
            {\small \bf IL } \\
            {\small Lisa flew 256 miles at 32 miles per hour. How long did Lisa fly? } \\ [.05in]
            {\small \em Translation produced: } \\
            {\small 256 32 $/$ } \\ [.1in]
            {\small \bf MAWPS } \\
            {\small Debby's class is going on a field trip to the zoo. If each van can hold 4 people and there are 2 students and 6 adults going, how many vans will they need? } \\ [.05in]
            {\small \em Translation produced: } \\
            {\small 2 6 + 4 $/$ } \\
            \hline
            \end{tabular}
        \end{figure}
        
    \subsubsection{Experiment 1 Analysis}
        All of the network configurations used were very successful for our task.
        The prefix representation overall provides the most stable network performance.
        We note that while the combined averages of the prefix models outperformed postfix, the postfix representation Transformer produced the highest average for a single model.
        The type 2 postfix Transformer received the highest testing average of 87.2\%.
        To highlight the capability of our most successful model (type 2 postfix Transformer), we present some outputs of the network in Figure \ref{figure:successfulTranslations}.
        
        The models respect the syntax of math expressions, even when incorrect.
        For most questions, our translators were able to determine operators based solely on the context of language.
        
        Table \ref{table:Experiment1Results} provides detailed results of Experiment 1.
        The numbers are absolute accuracies, i.e., they correspond to cases where the arithmetic expression generated is 100\% correct, leading to the correct numeric answer.
        Results by \cite{wang2018mathdqn,hosseini2014learning,roy2015reasoning,robaidek2018data} are sparse but indicate the scale of success compared to recent past approaches.
        Prefix, postfix, and infix representations in Table \ref{table:Experiment1Results} show that network capabilities are changed by how teachable the target data is.
        
        While our networks fell short of Wang, et al. AI2 testing accuracy \cite{wang2018mathdqn}, we present state-of-the-art results for the remaining three datasets in Table \ref{table:Experiment1Results}.
        The AI2 dataset is tricky because its questions contain numeric values that are extraneous or irrelevant to the actual computation, whereas the other datasets have only relevant numeric values. 
        Following these observations, we continue to more involved experiments with only the type 2 postfix Transformer.
        The next sections will introduce our preprocessing methods.
        Note that we start from scratch in our training for all experiments following this section.

    \section{Experiment 2: Preprocessing for Improved Results}
        We use various additional preprocessing methods to improve the training and testing of MWP data.
        One goal of this section is to improve the notably low performance on the AI2 tests.
        We introduce eight techniques for improvement and report our results as an average of 3 separate training and testing sessions.
        These techniques are also tested together in some cases to observe their combined effects.

        \subsection{Preprocessing Algorithms}
            We take note of previous pitfalls that have prevented neural approaches from applying to general MWP question answering.
            To improve English to equation translation further, we apply some transformation processes before the training step in our neural pipeline.
            First, we optionally remove all stop words in the questions.
            Then, again optionally, the words are transformed into a lemma to prevent easy mistakes caused by plurals.
            These simple transformations can be applied in both training and testing and require only base language knowledge.
            
            We also try minimalistic manipulation approaches in preprocessing and analyze the results.
            While in most cases, the tagged numbers are relevant and necessary when translating; in some cases, there are tagged numbers that do not appear in equations.
            To avoid this, we attempt three different methods of preprocessing: Selective Tagging, Exclusive Tagging, and Label-Selective Tagging.

            When applying Selective Tagging, we iterate through the words in each question and only replace numbers appearing in the equation with a tag representation (e.g., $\rm \langle j \rangle$).
            In this method, we leave the original numeric values in each sentence, which have do not collect importance in network translations.
            Similarly, we optionally apply Exclusive Tagging, which is nearly identical to Selective Tagging, but instead of leaving the numbers not appearing in the equation, we remove them.
            These two preprocessing algorithms prevent irrelevant numbers from mistakenly being learned as relevant.
            These methods are only applicable to training.

            It is common in MWPs to provide a label or indication of what a number represents in a question.
            For example, if we observe the statement ``George has 2 peaches and 4 apples. Lauren gives George 5 of her peaches. How many peaches does George have?" we only need to know quantifiers for the label ``peach."
            We consider ``peaches" and ``apples" as labels for the number tags.
            For the most basic interpretation of this series of mathematical prose, we know that we are supposed to use the number of peaches that George and Lauren have to formulate an expression to solve the MWP.
            Thus, determining the correct labels or tags for the numbers in an MWP is likely to be helpful.
            Similar to Selective Tagging and Exclusive Tagging, we avoid tagging irrelevant numbers when applying Label-Selective Tagging.
            Here, the quantity we need to ignore for a reliable translation is: ``4 apples.''
            This is as if we are associating each number with the appropriate unit notation, like ``kilogram" or ``meter."

            The Label-Selective Tagging method iterates through every word in an MWP question.
            We first count the occurrence of words in the question sentences and create an ordered list of all terms.
            In our example, we note that the word ``peaches" occurs three times in the sentence.
            Compared to the word ``apples" (occurring only once), we reduce our search for numbers to only the most common terms in each question.
            We impose a check to verify that the most common term appears in the sentence ending in a question mark, and if it is not, the Label-Selective Tagging fails and produces tags for all numbers.

            If we can identify the label reliably, we then look at each number.
            We assume that labels for quantities are within a window of three (either before the number or after).
            When we detect a number, we then look at four words before the number and four words after the number, and before any punctuation for the most common word we have previously identified.
            If we find the assumed label, we tag the number, indicating it is relevant.
            Otherwise, we leave the word as a number, which indicates that it is irrelevant.
            This method can be applied in training and testing equally because we do not require any knowledge about our target translation equation.
            We could have performed noun phrase chunking and some additional processing to identify nouns and their numeric quantifiers.
            However, our heuristic method works very well.

            In addition to restrictive tagging methods, we also try replacing all words with an equivalent part-of-speech denotation.
            The noun, ``George," will appear as ``NN,'' for example.
            There are two ways we employ this method.
            The first substitutes the word with its part-of-speech counterpart, and the second adds on the part-of-speech tag like ``(NN George),'' for each word in the sentence.
            These two algorithms can be applied both in training and testing since the part-of-speech tag comes from the underlying English language.
            
            We also try reordering the sentences in each MWP.
            For each of the questions, we sort the sentences in random order, not requiring the question within the MWP to appear last, as it typically does.
            The situational context is not linear in most cases when applying this transformation.
            We check to see if the network relies on the linear nature of MWP information to be successful.
            
            The eight algorithms presented are applied solo and in combination, when applicable.
            For a summary of the preprocessing algorithms, refer to Table \ref{table:PreprocessingAlgorithms}.
            \begin{table}[!h]
                \caption{Summary of Algorithms.}
                \label{table:PreprocessingAlgorithms}
                \begin{center}
                \begin{tabular}{p{0.59\linewidth}p{0.3\linewidth}}
                \hline
                {\small \bf Name} & {\small \bf Abbreviation} \\
                \hline
                {\small Remove Stop Words } & {\small SW } \\
                {\small Lemmatize } & {\small L }\\
                {\small Selective Tagging } & {\small ST }\\
                {\small Label-Selective Tagging } & {\small LST }\\
                {\small Exclusive Tagging } & {\small ET }\\
                {\small Part of Speech } & {\small POS }\\
                {\small Part of Speech w/ Words } & {\small WPOS }\\
                {\small Sentence Reordering } & {\small R }\\
                \hline
                \end{tabular}
                \end{center}
            \end{table}
            \begin{table*}[!ht]
                \caption{Test results for Experiment 2 (* denotes averages on present values only).
                Rows after the top 5 indicate type 2 postfix Transformer results. }
                \label{table:Experiment2Results}
                \centering
                \begin{tabular}{p{0.29\linewidth}p{0.11\linewidth}p{0.11\linewidth}p{0.11\linewidth}p{0.12\linewidth}p{0.1\linewidth}}
                \hline
                {\small \bf Preprocessing Method} & {\small \bf AI2 } & {\small \bf CC } & {\small \bf IL } & {\small \bf MAWPS } & {\small \bf Average } \\
                \hline
                {\small \cite{hosseini2014learning} } & {\small 77.7 } & {\small -- } & {\small -- } & {\small -- } & {\small $^*$77.7 } \\
                {\small \cite{kushman2014learning} } & {\small 64.0 } & {\small 73.7 } & {\small 2.3 } & {\small -- } & {\small $^*$46.7 } \\
                {\small \cite{roy2015reasoning} } & {\small -- } & {\small -- } & {\small 52.7 } & {\small -- } & {\small $^*$52.7 } \\
                {\small \cite{robaidek2018data} } & {\small -- } & {\small -- } & {\small -- } & {\small 62.8 } & {\small $^*$62.8 } \\
                {\small \cite{wang2018mathdqn} } & {\small 78.5 } & {\small 75.5 } & {\small 73.3 } & {\small -- } & {\small $^*$75.4 } \\ [.05in]
                {\small \em Type 2 Postfix-Transformer } & & & & & \\
                {\small None } & {\small 77.2 } & {\small 94.4 } & {\small 94.1 } & {\small 83.1 } & {\small 87.2 } \\
                {\small SW } & {\small 73.7 } & {\small \bf{100.0} } & {\small \bf{100.0} } & {\small \bf 94.0 } & {\small 91.9 } \\
                {\small L } & {\small 63.2 } & {\small 86.7 } & {\small 80.9 } & {\small 68.1 } & {\small 74.7 } \\
                {\small SW + L } & {\small 61.4 } & {\small 60.0 } & {\small 57.1 } & {\small 66.4 } & {\small 61.2 } \\
                {\small ST } & {\small 80.7 } & {\small 93.3 } & {\small \bf{100.0} } & {\small 84.3 } & {\small 89.6 } \\
                {\small SW + ST } & {\small 71.9 } & {\small 93.3 } & {\small \bf{100.0} } & {\small 83.5 } & {\small 87.2 } \\
                {\small SW + L + ST } & {\small 50.9 } & {\small 63.3 } & {\small 48.8 } & {\small 64.1 } & {\small 56.8 } \\
                {\small LST } & {\small \bf{82.5} } & {\small \bf{100.0} } & {\small \bf{100.0} } & {\small 93.7 } & {\small \bf 94.0 } \\
                {\small SW + LST } & {\small 70.2 } & {\small \bf{100.0} } & {\small \bf{100.0} } & {\small 92.6 } & {\small 90.7 } \\
                {\small SW + L + LST } & {\small 54.4 } & {\small 71.1 } & {\small 50.0 } & {\small 68.1 } & {\small 60.9 } \\
                {\small ET } & {\small 80.7 } & {\small 93.3 } & {\small \bf{100.0} } & {\small 84.0 } & {\small 89.5 } \\
                {\small SW + ET } & {\small 70.2 } & {\small 93.3 } & {\small \bf{100.0} } & {\small 84.0 } & {\small 86.9 } \\
                {\small SW + L + ET } & {\small 59.6 } & {\small 63.3 } & {\small 53.6 } & {\small 66.1 } & {\small 60.7 } \\
                {\small POS } & {\small 79.0 } & {\small \bf{100.0} } & {\small 73.8 } & {\small 91.5 } & {\small 86.1 } \\
                {\small WPOS } & {\small 40.4 } & {\small 0.0 } & {\small 84.5 } & {\small 53.0 } & {\small 44.5 } \\
                {\small R } & {\small 38.0 } & {\small 59.6 } & {\small 58.7 } & {\small 42.3 } & {\small 49.6 } \\
                \hline
                \end{tabular}
            \end{table*}

        \subsubsection{Experiment 2 Results and Analysis}
            We present the results of Experiment 2 in Table \ref{table:Experiment2Results}.
            From the results in Table \ref{table:Experiment2Results}, we see that Label-Selective Tagging is very successful in improving the translation quality of MWPs.
            Other methods, such as removing stop words, improve accuracy due to the reduction in the necessary vocabulary, but fail to outperform LST for three of the four datasets.
            Using a frequency measure of terms to determine number relevancy is a simple addition to the network training pipeline and significantly outperforms our standalone Transformer network base.

            The LST algorithm was successful for two reasons.
            The first reason LST is more realistic in this application is its applicability to inference time translations.
            Because the method relies only on each question's vocabulary, there are no restrictions on usability.
            This method reliably produces the same results in the evaluation as in training, which is a unique characteristic of only a subset of the preprocessing algorithms tested.
            The second reason LST is better for our purpose is that it prevents unnecessary learning of irrelevant numbers in the questions.
            One challenge in the AI2 dataset is that some numbers present in the questions are irrelevant to the intended translation.
            Without some preprocessing, we see that our network sometimes struggles to determine irrelevancy for a given number.
            LST also reduces compute time for other areas of the data pipeline.
            Only a fraction of the numbers in some questions need to be tagged for the network, which produces less stress on our number tagging process.
            Still some common operator inference mistakes were made while using LST, shown in Figure \ref{figure:unsuccessfulTranslation}.

            The removal of stop words is a common practice in language classification tasks and was somewhat successful in translation tasks.
            We see an improvement on all datasets, suggesting that stop words are mostly ignored in successful translations and paid attention to more when the network makes mistakes.
            There is a significant drop in reliability when we transform words into their base lemmas.
            Likely, the cause of this drop is the loss of information by imposing two filtering techniques at once.
            
            \begin{figure}
                \caption{Example of an unsuccessful translation using type 2 postfix Transformer and LST.}
                \label{figure:unsuccessfulTranslation}
                \centering
                \begin{tabular}{p{0.9\linewidth}}
                \hline
                {\small \bf MAWPS } \\
                {\small There were 73 bales of hay in the barn. Jason stacked bales in the barn today. There are now 96 bales of hay in the barn. How many bales did he store in the barn? } \\ [.05in]
                {\small \em Translation produced: } \\
                {\small 96 73 - } \\ [.1in]
                {\small \em Expected translation: } \\
                {\small 73 96 + } \\
                \hline
                \end{tabular}
            \end{figure}
            Along with the successes of the tested preprocessing came some disappointing results.
            Raw part-of-speech tagging produces slightly improved results from the base model, but including the words and the corresponding part-of-speech denotations fail in our application.

            Reordering of the question sentences produced significantly worse results.
	        The accuracy difference mostly comes from the random position of the question, sometimes appearing with no context to the transaction.
            Some form of limited sentence reordering may improve the results, but likely not the degree of success of the other methods.

            By incorporating simple preprocessing techniques, we grow the generality of the Transformer architecture to this sequence-to-sequence task.

    \subsubsection{Overall Results}
        Table \ref{table:Experiment2Results} shows that the Transformer architecture with simple preprocessing outperforms previous state-of-the-art results in all four tested datasets.
        While the Transformer architecture has been well-proven in other tasks, we show that applying the attention schema here improves MWP solving.

        With our work, we show that relatively small networks are more stable for our task.
        Postfix and prefix both are a better choice for training neural networks, with the implication that infix can be re-derived if it is preferred by the user of a solver system.
        The use of alternative mathematical representation contributes greatly to the success of our translations.

    \section{Conclusions and Future Work}
        \balance
        In this paper, we have shown that the use of Transformer networks improves automatic math word problem-solving.
        We have also shown that postfix target expressions perform better than the other two expression formats.
        Our improvements are well-motivated but straightforward and easy to use, demonstrating that the well-acclaimed Transformer architecture for language processing can handle MWPs well, obviating the need to build specialized neural architectures for this task.

        In the future, we wish to work with more complex MWP datasets.
        Our datasets contain basic arithmetic expressions of +, -, *, and /, and only up to 3 of them.
        For example, datasets such as Dolphin18k \cite{huang2016well}, consisting of web-answered questions from Yahoo! Answers, require a wider variety of language to be understood by the system.

        We wish to use other architectures stemming from the base Transformer to maximize the accuracy of the system.
        For our experiments, we use the 2017 variation of the Transformer \cite{vaswani2017attention}, to show that generally applicable neural architectures work well for this task.
	    With that said, we also note the importance of a strategically designed architecture to improve our results.

        To further the interest in automatic solving of math word problems, we have released all of the code used on GitHub.\footnote{\url{https://github.com/kadengriffith/MWP-Automatic-Solver}}

    \bibliography{Griffith_Kalita}
    \bibliographystyle{acl_natbib}
\end{document}